\documentclass{article}

\title{An Overview of General Performance Metrics of Binary Classifier Systems}
\author{Sebastian Raschka\\ \texttt{se.raschka@gmail.com}} 
\date{\today}

\usepackage{cite}
\usepackage{graphicx}	
\usepackage{hyperref}
\usepackage{titlesec}
\titlespacing*{\section}
{0pt}{5.5ex plus 1ex minus .2ex}{4.3ex plus .2ex}

\usepackage{fancyhdr}
\usepackage{array}
\usepackage{multirow}

\newcommand\MyBox[2]{
  \fbox{\lower0.75cm
    \vbox to 1.7cm{\vfil
      \hbox to 1.7cm{\hfil\parbox{1.4cm}{#1\\#2}\hfil}
      \vfil}%
  }%
}

\begin{document} 

\maketitle 

\noindent  The purpose of this document is to provide a brief overview of different metrics and terminology that is used to measure the performance of binary classification systems.

\tableofcontents

\newpage

\section{Confusion matrix}

The \emph{confusion matrix} (or \emph {error matrix}) is one way to summarize the performance of a classifier for binary classification tasks. This square matrix consists of columns and rows that list the number of instances as absolute or relative "actual class" vs. "predicted class" ratios.

\noindent Let $P$ be the label of class 1 and $N$ be the label of a second class or the label of all classes that are \emph{not class 1} in a multi-class setting.

\noindent
\renewcommand\arraystretch{1.5}
\setlength\tabcolsep{0pt}
\begin{tabular}{c >{\bfseries}r @{\hspace{0.7em}}c @{\hspace{0.4em}}c @{\hspace{0.7em}}l}
  \multirow{10}{*}{\parbox{1.1cm}{\bfseries\raggedleft Actual\\ Class}} 
    & \multicolumn{3}{c}{\bfseries Predicted class} & \\
  & & \bfseries $P$  & \bfseries $N$ & \\
  & $P$ & \MyBox{True}{Positives (TP)} & \MyBox{False}{Negatives (FN)}  &\\[2.4em]
  & $N$ & \MyBox{False}{Positives (FP)} & \MyBox{True}{Negatives (TN)} & \\
\end{tabular}

\vspace{1cm}

\noindent 
The following equations are based on \emph{An introduction to ROC analysis} by Tom Fawcett \cite{fawcett2006introduction}.

\section{Prediction Error and Accuracy}

Both the prediction \emph{error (ERR)} and \emph{accuracy (ACC)} provide general information about how many samples are misclassified. The \emph{error} can be understood as the sum of all false predictions divided by the number of total predications, and the the accuracy is calculated as the sum of correct predictions divided by the total number of predictions, respectively. 

\begin{equation} ERR = \frac{FP + FN}{FP+ FN + TP + TN} = 1-ACC\end{equation}

\begin{equation} ACC = \frac{TP + TN}{FP+ FN + TP + TN} = 1-ERR\end{equation}

\section{False and True Positive Rates}

The \emph{True Positive Rate (TPR)} and \emph{False Positive Rate (FPR)} are performance metrics that are especially useful for imbalanced class problems. In \emph{Spam classification}, for example, we are of course primarily interested in the detection and filtering out of \emph{spam}. However, it is also important to decrease the number of messages that were incorrectly classified as \emph{spam} (\emph{False Positives}): A situation where a person misses an important message is considered as "worse" than a situation where a person ends up with a few \emph{spam} messages in his e-mail inbox. In contrast to the \emph{FPR}, the \emph{True Positive Rate} provides useful information about the fraction of \emph{positive} (or \emph{relevant}) samples that were correctly identified out of the total pool of \emph{Positives}.

\begin{equation} FPR = \frac{FP}{N} =  \frac{FP}{FP + TN} \end{equation}

\begin{equation} TPR = \frac{TP}{P} =  \frac{TP}{FN + TP} \end{equation}

\section{Precision, Recall, and the $F_1$-Score}

\emph{Precision (PRE)} and \emph{Recall (REC)} are metrics that are more commonly used in \emph{Information Technology} and related to the \emph {False} and \emph{True Prositive Rates}. In fact, \emph{Recall} is synonymous to the \emph{True Positive Rate} and also sometimes called \emph{Sensitivity}. The F$_1$-Score can be understood as a combination of both \emph{Precision} and \emph{Recall} \cite{goutte2005probabilistic}.

\emph{Precision (PRE)} and \emph{Recall (REC)} are metrics that are more commonly used in \emph{Information Technology} and related to the \emph {False} and \emph{True Prositive Rates}. In fact, \emph{Recall} is synonymous to the \emph{True Positive Rate} and is sometimes also called \emph{Sensitivity}. The $F_1$-Score can be understood as a combination of both \emph{Precision} and \emph{Recall} \cite{goutte2005probabilistic}.

\begin{equation} PRE = \frac{TP}{TP + FP} \end{equation}

\begin{equation} REC = TPR = \frac{TP}{P} =  \frac{TP}{FN + TP} \end{equation}

\begin{equation} F_1 = 2 \cdot \frac{PRE \cdot REC}{PRE + REC}\end{equation}

\section{Sensitivity and Specificity}

\emph{Sensitivity (SEN)} is synonymous to \emph{Recall} and the \emph{True Positive Rate} whereas \emph{Specificity (SPC)} is synonymous to the \emph{True Negative Rate} --- Sensitivity measures the recovery rate of the \emph{Positives} and complimentary, the Specificity  measures the recovery rate of the \emph{Negatives}.

\begin{equation} SEN = TPR = REC = \frac{TP}{P} =  \frac{TP}{FN + TP} \end{equation}
\begin{equation} SPC = TNR =\frac{TN}{N} =  \frac{TN}{FP + TN} \end{equation}

\section{Matthews correlation coefficient}

\emph{Matthews correlation coefficient (MCC)} was first formulated by Brian W. Matthews \cite{matthews1975comparison} in 1975 to assess the performance of protein secondary structure predictions. The MCC can be understood as a specific case of a linear correlation coefficient (\emph{Pearson r}) for a binary classification setting and is considered as especially useful in unbalanced class settings.
The previous metrics take values in the range between 0 (worst) and 1 (best), whereas the MCC is bounded between the range 1 (perfect correlation between ground truth and predicted outcome) and -1 (inverse or negative correlation) --- a value of 0 denotes a random prediction. 

\begin{equation} MCC = \frac{ TP \cdot TN - FP \cdot FN } {\sqrt{ (TP + FP) ( TP + FN ) ( TN + FP ) ( TN + FN ) } } \end{equation}

\section{Receiver Operator Characteristic (ROC)}

\emph{Receiver Operator Characteristics (ROC) graphs} are useful tools to select classification models based on their performance with respect to the \emph{False Positive} and \emph{True Positive} rates.

The diagonal of a ROC graph can be interpreted as \emph{random guessing} and classification models that fall below the diagonal are considered as worse than random guessing. A perfect classifier would fall into the top left corner of the graph with a \emph{True Positive Rate} of 1 and a  \emph{False Positive Rate} of 0.

\begin{figure}[h!]
    \centering
    \includegraphics[width=0.8\textwidth]{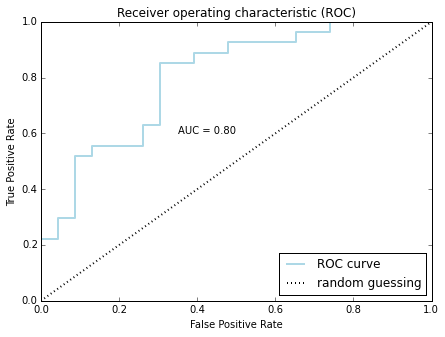}
    \caption{Example of a Receiver Operating Characteristic. This plot was created using the Python \href{http://scikit-learn.org/stable/auto_examples/plot_roc.html}{scikit-learn} machine learning library.}

\end{figure}

\noindent The ROC \emph{curve} can be computed by shifting the decision threshold of a classifier (e.g., the posterior probabilities of a naive Bayes classifier). Based on the  ROC \emph{curve}, the so-called \emph{Area Under the Curve (AUC)} can be calculated to characterize the performance of a classification model.

\newpage

\bibliography{./bib}{}

\begin{thebibliography}{1}

\bibitem{fawcett2006introduction}
Tom Fawcett.
\newblock An introduction to roc analysis.
\newblock {\em Pattern recognition letters}, 27(8):861--874, 2006.

\bibitem{goutte2005probabilistic}
Cyril Goutte and Eric Gaussier.
\newblock A probabilistic interpretation of precision, recall and f-score, with
  implication for evaluation.
\newblock In {\em Advances in Information Retrieval}, pages 345--359. Springer,
  2005.

\bibitem{matthews1975comparison}
Brian~W Matthews.
\newblock Comparison of the predicted and observed secondary structure of t4
  phage lysozyme.
\newblock {\em Biochimica et Biophysica Acta (BBA)-Protein Structure},
  405(2):442--451, 1975.

\end{thebibliography}
\bibliographystyle{plain}

\end{document}